\documentclass[final,5p,times,twocolumn]{elsarticle}

\usepackage{graphicx}
\usepackage[utf8]{inputenc}
\usepackage{graphics, subfig} % for pdf, bitmapped graphics files
\usepackage{epsfig} % for postscript graphics files
\usepackage{mathptmx} % assumes new font selection scheme installed
\usepackage{amsmath} % assumes amsmath package installed
\usepackage{amsfonts}
\usepackage{amssymb}  % assumes amsmath package installed
\usepackage{lineno,hyperref}
\modulolinenumbers[5]
\usepackage{subfig} % subfigures
\usepackage{forest} %taxonomy forests
\usepackage[euler]{textgreek} %greek letters in normal text
\usepackage{mathtools}
\DeclarePairedDelimiterX{\norm}[1]{\lVert}{\rVert}{#1}
\usepackage{multirow}

\journal{Information Fusion}

\begin{document}

\begin{frontmatter}

\newtheorem{sample}{Teorema}
\newtheorem{definition}[sample]{Definition}

\title{Non-IID data and Continual Learning processes in\\ Federated Learning: A long road ahead}

%% Authors:
\author[citius]{Marcos F. Criado\corref{mycorrespondingauthor}}
\cortext[mycorrespondingauthor]{Corresponding author}
\ead{marcos.criado@usc.es}

\author[citius]{Fernando E. Casado}\ead{fernando.estevez.casado@usc.es}

\author[citius]{Roberto Iglesias}\ead{roberto.iglesias.rodriguez@usc.es}

\author[fic]{Carlos V. Regueiro}\ead{carlos.vazquez.regueiro@udc.es}

\author[citius]{Sen\'en Barro}\ead{senen.barro@usc.es}

\address[citius]{CiTIUS (Centro Singular de Investigaci\'on en Tecnolox\'ias Intelixentes), Universidade de Santiago de Compostela, 15782 Santiago de Compostela, Spain }

\address[fic]{CITIC, Computer Architecture Group, Universidade da Coru\~na, 15071 A Coru\~na, Spain}

%% Abstract:
\begin{abstract}

Federated Learning is a novel framework that allows multiple devices or institutions to train a machine learning model collaboratively while preserving their data private. This decentralized approach is prone to suffer the consequences of data statistical heterogeneity, both across the different entities and over time, which may lead to a lack of convergence. To avoid such issues, different methods have been proposed in the past few years. However, data may be heterogeneous in lots of different ways, and current proposals do not always determine the kind of heterogeneity they are considering. In this work, we formally classify data statistical heterogeneity and review the most remarkable learning strategies that are able to face it. At the same time, we introduce approaches from other machine learning frameworks, such as Continual Learning, that also deal with data heterogeneity and could be easily adapted to the Federated Learning settings. 

\end{abstract}

% Keywords:
\begin{keyword}
federated learning \sep data heterogeneity \sep non-IID data \sep concept drift \sep distributed learning \sep continual learning.
\end{keyword}

\end{frontmatter}

%\linenumbers

\section{Introduction}
\label{sec:introduction}

Machine Learning (ML) consists of the study of mathematical algorithms that improve automatically through experience with the use of data. Traditionally, data used for training ML algorithms are gathered in a centralized dataset, and the process of training can access each data sample at any time. However, in addition to databases, nowadays we live in a society of devices where the main primary computing machines for people in their daily life are smartphones and tablets, equipped with cutting-edge sensors, and computing and communication capabilities. Those devices collect useful data prone to be used for training personalized algorithms that simplify their daily usage. In this context, the quantity of data recorded in just one device may not be sufficient for obtaining an accurate model to perform the desired task. To solve this matter, in the past few years a new paradigm of ML, Federated Learning (FL)~\cite{konevcny2015federated,mcmahan2017communication,lim2020federated}, was developed. This new learning strategy is based on the idea of training a joint model using data from a multitude of coordinated devices in a decentralized way, and has achieved impressive results.

Several problems arise when trying to train a model under these circumstances. For instance, each device has its own processing and storage capacities, which leads to differences in the time needed to perform the training stage~\cite{li2020federated}. In this paper, we will focus on discussing the statistical variability attached to the use of a myriad of different sets of data, with samples collected in distinct situations. This heterogeneity in the samples is also referred to as \emph{non-IID data}~\cite{zhao2018federated}, and is one of the main difficulties encountered in the federated learning process.  Assuming data is Independent and Identically Distributed (IID) to avoid some complications, as many works do, is not a good option. Different devices may collect very distinct samples, or even contradictory ones. We will analyze and compare the strategies established so far to face these kinds of issues. 

One other assumption in standard ML is that the whole set of samples is available from the beginning of the training stage. However, in realistic issues, it is frequent to collect data progressively, during several days, or weeks, depending on the performed task. For this reason, Continual Learning (CL)~\cite{parisi2019continual} research gains a lot of attention, since it addresses the difficulties of training a model gradually using real-time collected data. In particular, it faces variations in data as time passes. This phenomenon is known as concept drift~\cite{gama2014survey}.

In this work, we present some of the possible scenarios that can arise when trying to solve a real problem applying FL, and the difficulties that need to be faced. We classify those scenarios attending to the statistical heterogeneity of data, combining the federated and continual settings to visualize the whole problem scenarios, and we present a collection of the most remarkable techniques that have been studied to deal with some of those issues. We also notice that some real situations have not been considered nor handled so far, and they should be taken into account~\cite{casado2021concept,usmanova2021distillation,park2021tackling}.

The rest of this paper is organized as follows: Section~\ref{sec:federated} reviews the state-of-the-art techniques for Federated Learning. In Section~\ref{sec:federated-non-iid}, we present the definition and classification of non-IID data in a federated environment, and we also discuss the different strategies to deal with it. In Section~\ref{sec:continual}, we introduce the Continual Learning framework and the multiple ways data can evolve over time. In Section~\ref{sec:adressing-non-iid-data}, we combine the different situations of heterogeneous data to show all of the possible scenarios. In addition, we discuss the strategies used to train a model under concept drift that are close to the federated learning framework. Finally, Section~\ref{sec:challenges} gathers our main conclusions and unsolved challenges.

\section{Overview on Federated Learning}
\label{sec:federated}

Federated Learning (FL)~\cite{konevcny2015federated,mcmahan2017communication,lim2020federated} is a ML framework used to train an algorithm across multiple decentralized edge devices, where each of them holds its own local data samples. This approach arose in 2016. Federated learning enables multiple clients to build a robust, joint ML model without sharing data, thus allowing to address essential issues such as data privacy and security. The general idea of this technique consists of a loop of local learning stages in the devices, and global parameter aggregations in a central server in the cloud; in such a way that the only shared information are those parameters. There is a model shared by the clients to perform their local training, which is usually a Deep Neural Network (DNN). The most popular federated algorithm is Federated Averaging (FedAvg)~\cite{mcmahan2016federated}.

There are some key components in standard federated settings. The parameters $w \in \mathit{W}$ of the model are initialized on the server, so that every participant $i \in \{1,\ldots,n\} = \mathit{N}$ starts the training stage at a common point. This is crucial for convergence purposes. In each federated round $r$, a random subset of clients, $i \in \mathit{N}^{r}~\subseteq~\{1,2, \dots, n\}$, of size $|\mathit{N}^{r}|$, is selected. Those devices receive the current parameter set from the server, $w_G^r$, perform stochastic gradient descent (SGD) on their local datasets, $\mathit{D}_i \subset \mathit{X} \times \mathit{Y}$, and send back the updated parameters, $w_i^{r+1}$. Federated learning is typically deployed in a supervised setting, where the data samples $(x,y) \in \mathit{D}_i$ consist of a pair of input data $x \in \mathit{X}$ and its correspondent output $y \in \mathit{Y}$. After performing local training, the local results are aggregated in the server to update the model parameters and start the next round. In the particular case of FedAvg, the local parameters from each client are aggregated applying a weighted mean, to balance the different dataset sizes:
\begin{equation*}
    \mathbf{w}_G^{r} \leftarrow \sum_{i=1}^{N} \frac{M_i}{M} \mathbf{w}_i^{r},
\end{equation*}
where $M$ is the total number of data instances at each round, and $M_i$ is the number of instances from client $i$. After that, a new training round starts, a random subset of clients is selected and the updated model is sent back to those clients.

It is important to notice that most of the works in the literature perform supervised FL, although there are existing works that consider unsupervised settings. The case of reinforcement learning has been little studied so far, and there is not a significant volume of works in that area. Consequently, in this review we will only consider supervised and unsupervised strategies.

Another key point in the federated setting is privacy and data security. The distributed spirit of this framework enhances the possibility of training a shared model without actually sharing any piece of information. This is not only desirable but also mandatory. In fact, over the last few years, several governments around the world have implemented data privacy legislation to protect consumers, limiting their data sending and storage only to what is consented by them, and absolutely necessary for processing. Some examples of this are the European Commission's General Data Protection Regulation (GDPR)~\cite{custers2019eu} or the Consumer Privacy Bill of Rights in the US~\cite{gaff2014privacy}. In addition, a lot of research has been done concerning information sensibility and model threats~\cite{lyu2020threats}. For instance, guarantees about the lack of information of the gradient and the neural network weights have been widely studied to preserve client privacy~\cite{truex2019hybrid}. Even so, the general thought is the Federated Learning framework alone is not enough to keep data privacy~\cite{wang2019beyond}, and for this reason there are different strategies to obscure the information, such as Homomorphic Encryption~\cite{naehrig2011can,aslett2015review}, Differential Privacy~\cite{wei2020federated,geyer2017differentially} or Secure Multi-Party Computation~\cite{zhao2019secure,liu2020secure}.

However, the training framework we just presented is very recent and innovative, hence presents some challenges that have not been fully addressed yet. For instance, participants selected to train may drop out, or be incapable of performing a local update for the model, due to factors like poor connection or lack of battery. To avoid these issues, there are two options. On the one hand, there are some specific strategies, named Asynchronous FL~\cite{casado2020federated,chen2020asynchronous,li2021stragglers}, designed to deal with these inconveniences. devices that check in to the server are required to be plugged in, on a proper connection, and idle, in order to avoid impacting the user of the device. Concerning the convergence of the algorithm, the federated framework presents some shortages. There are no guarantees that training a model under limited communications between local devices and the central server and little computation resources in the devices would lead to a successful result. In fact, there are studies that expose the possible problem of having adversarial participants sending miss-labeled, data-poisoned updates to prevent the model convergence \cite{rodriguez2020dynamic}. Apart from adversarial attacks, theoretical and experimental analysis have been carried out on gradient-descent approaches assuming a convex objective loss function ~\cite{wang2019adaptive,dinh2020federated}.

Despite being a novel paradigm of ML, FL has attracted a lot of attention. Given the vast amount of research done in the field in the past few years, it became important to keep track of the most remarkable advances. For that reason, it is possible to find reviews on FL that focus on specific issues and show the advances achieved, as well as the remaining open problems. For instance, one of the biggest challenges we already discussed is keeping the clients privacy. This problem is widely studied in~\cite{kairouz2019advances}, where the authors focus not only on privacy guarantees of the federated settings but also on preventing malicious attacks conducted to steal pieces of information. Another interesting review is~\cite{li2020federated}, which concentrates its attention on the problem of communications and the different approaches conceived to perform it. Both of these works also acknowledge the problem of statistical heterogeneity of the data, showing their concerns and exposing the difficulties of working with it. However, little attention is paid to the actual methods that face this challenge, their advantages and their drawbacks. In this paper, we will focus on those matters.

Statistical heterogeneity of the data is a major issue that needs to be faced in order to construct and deploy a FL model. A bunch of devices training over different local datasets may produce updates in a wide range, thus leading to an undesired result after aggregation, or even worse, impeding the model to converge at all. To prevent these obstacles a common assumption in decentralized learning is considering that the data of the different participants is Independent and Identically Distributed (IID)~\cite{lim2020federated,zhao2018federated,kairouz2019advances}. This means that data collected by different participants does not present significant differences. Nonetheless, in most real-life problems and situations this assumption is not satisfied: each client acts in a particular way, thus collects biased data which differs from the one collected by another participant. Further, clients may be interested in applying the global model obtained to predict information in different scenarios. All of these possibilities are gathered under the equivalent concepts of \emph{non-IID data} or \emph{data heterogeneity}, which results ambiguous. Also, in real-world problems it is important to account for another source of heterogeneity: devices constantly extract new data from their environment, so it is crucial to implement some Continual Learning strategy to adapt the present model to the current data samples~\cite{usmanova2021distillation,park2021tackling}.

There is another difficulty when evaluating heterogeneous data. To test if a method is able to provide the clients accurate models, a dataset that reflects the heterogeneity they present is required.  Currently, there are no specific sets of data designed to evaluate the goodness of a model that tries to face non-IID realistic problems. Most of the works we present throughout this paper perform their experimental results using a federated benchmark dataset, and modifying it to obtain the required heterogeneity. An example of this is the MNIST dataset~\cite{lecun1998gradient}. Lots of works use this dataset in their experiments, but in some works, images are rotated, different types of noise are added to the samples (domain shift), or the labels of two classes are exchanged with a fixed proportion (behaviour changes). In addition, some modifications had such an impact that they were preserved as distinct datasets, such as MNIST-M~\cite{ganin2016domain}. In the end, each strategy is experimentally tested with a dataset customized in a unique way, distinct from what other works with the same objective do. This is a huge problem as it stands in the way of properly comparing the different methods and analysing which one provides a better result. Establishing at least one common benchmark dataset that represents several kinds of heterogeneous data should be a priority.

\section{Non-IID data in Federated Learning}
\label{sec:federated-non-iid}

Lots of research has been done regarding the issue of dealing with non-IID data, specially in the context of Federated Learning, where it acquires great importance. In this paper, we will use the words `heterogeneous data' as a synonym for non-IID data. Existing works focus on both developing new techniques to tackle data heterogeneity~\cite{zhao2018federated,wang2020optimizing}, and proving the convergence of traditional FedAvg trained with non-IID data under some restrictive assumptions~\cite{sattler2019robust,li2019convergence}. However, in most cases, little specifications are made concerning the heterogeneity source of data, if made at all. 

Firstly, we are going to give a formal definition of \emph{Independent and Identically Distributed data} (IID data). For that, we need to settle the data probability distributions of the different data owners. Recall we denote a client by $i \in \mathit{N}$, with $\mathit{N} = \{1,\ldots,n\}$ being the set of all clients. We also denote each data sample as $d = (x,y) \in \mathit{X} \times \mathit{Y}$, where $x$ is the feature vector of the sample, and $y$ is the corresponding label, in supervised settings. Each client collects its own data $\mathit{D}_i$, and therefore it has a data probability distribution $\mathit{D}_i(x,y)$, where each data sample $(x,y)$ has probability $\mathit{P}_i(x,y)$. The overall data distribution is a weighted mean of these local data distributions:
\begin{equation} \label{globaldistribution}
\mathit{D}_G (x,y) = \sum_{i=1}^N \frac{M_i}{M} \cdot \mathit{D}_i(x,y)
\end{equation}
where $M_i$ is the amount of data collected by the $i$-th device, and $M = \sum_{i=1}^N M_i$ is the total number of data samples. Recall that we are working under standard FL assumptions, i.e., the local datasets are fully available from the beginning, so we do not have to deal with shifts in the distributions over time. For that reason, we do not use any temporal index.

\begin{definition}
Data is said to be IID if the probability belonging to a data sample does not vary as other samples are drawn, and every sample randomly selected can belong to any local dataset with the same probability. In mathematical terms, if we consider the union of the datasets $\mathit{D} = \bigcup_{i=1}^N \mathit{D}_i \subset \mathit{X} \times \mathit{Y}$, we claim $\mathit{D}$ is an IID dataset if, and only if,
    \begin{equation} \label{eq:iid_definition}
        \begin{cases}
         \mathit{P}_i((x,y),(x',y')) = \mathit{P}_i(x,y) \cdot \mathit{P}_i(x',y'), \\
        \mathit{D}_i(x,y) = \mathit{D}_j(x,y).
        \end{cases}
    \end{equation}
For all $ i,j \in \mathit{N}$, and for all $(x,y) , (x',y') \in \mathit{D}_i \cup \mathit{D}_j.$
\end{definition} 

This definition has one important consequence: Under IID data, all the clients distributions are equal to the global distribution:
\[ \mathit{D}_G(x,y) = \sum_{i=1}^N \frac{M_i}{M} \mathit{D}_i(x,y) = \mathit{D}_i(x,y) \sum_{i=1}^N \frac{M_i}{M} = \mathit{D}_i(x,y) \ \forall i \in \mathit{N}. \]
Reciprocally, we say data is non-IID if any of the two conditions given in Equation~(\ref{eq:iid_definition}) are not satisfied. However, this situation is quite less informative than the one from the IID data scenario, since we are unaware of which affirmation is not satisfied, where do the distributions differences lie, etc. 

Notice that a local dataset is always IID if the samples it holds are independent. In other words, in settings where the data is centralized or gathered together, data is always identically distributed, since there is only one device involved in training. The term \emph{non-IID} in machine learning implies the existence of various participants, or sets of data, and it is mostly used in the decentralized paradigm.

\subsection{Taxonomy of data heterogeneity} 
\label{sec:non-iid-classification}

Data can be non-IID for many reasons. For instance, there may exist a partition of the clients such that each group presents an IID dataset, but the mixture of them turns out to be non-IID. In fact, this is the reasoning behind the \emph{Group-level personalization} techniques developed in FL (Section~\ref{sec:grouplevelpersonal}). Another option would be that the participants local datasets present slightly different properties while sharing some others. This situation could be handled with \emph{Client-level personalization} strategies (Section~\ref{sec:clientlevelpersonal}). On the whole, the reason for the data to be non-IID is an important piece of information to decide the most convenient approach to face it. Hence, we want to dig deeper into the possible causes of disturbances in the joint probabilities. These causes can rely on multiple elements, and lead to unequal distributions among the clients~\cite{kairouz2019advances,li2021federated}. To characterize the possible causes of non-IID data, it happens to be more useful to think in terms of the probability density functions, $\mathit{P}(x,y)$ and $\mathit{P}_i(x,y)$, rather than distributions, since they can be factorized in two different ways:
\begin{equation} \label{eq:factorization}
    \mathit{P}(x,y) = \mathit{P}(x) \cdot \mathit{P}(y | x),
\end{equation}
\begin{equation} \label{eq:otherfactorization}
    \mathit{P}(x,y) = \mathit{P}(y) \cdot \mathit{P}(x | y).
\end{equation}
Given these factorizations, we can better distinguish which term represents the clients particularities. If we are dealing with clients who own data samples unbalanced over the possible classes, the difference between probabilities would lie in the terms $\mathit{P}(y)$ or $\mathit{P}(y | x)$. However, in this work we are going to focus on the factorization given by Equation~(\ref{eq:factorization}), since the conditional probability $\mathit{P}(x | y)$ in Equation~(\ref{eq:otherfactorization}) may seem controversial with the natural way of training a model, which consists of predicting $y$ based on $x$ and not the other way around. 

If the data probabilities belonging to $2$ participants, $\mathit{P}_i(x,y)$ and $\mathit{P}_j(x,y)$, are the same, then both factors would be equal: $\mathit{P}_i(x) = \mathit{P}_j(x)$ and $\mathit{P}_i(y |x) = \mathit{P}_j(y | x)$. Notice that every time we say two probability distributions are the same we mean they are alike in statistical terms, i.e., they cannot be recognized as different using a standard hypothesis testing. If, on the contrary, the joint probabilities are not the same, there are three possible scenarios according to the previous factorization:

\begin{itemize}
    \item[i)] $\mathit{P}_i(x) \neq \mathit{P}_j(x)$ and $\mathit{P}_i(y | x) = \mathit{P}_j(y | x)$. In this kind of situation, clients own data samples from different domains, but they share the same goal. This could be the case of, for example, participants collecting data for training an autonomous car. Some users may drive on the left and some others on the right, and they will face different circumstances. That will make the input space of the different participants skew. However, they gather data with one common objective, and they are expected to act similarly.
    \item[ii)] $\mathit{P}_i(x) = \mathit{P}_j(x)$ and $\mathit{P}_i(y |x) \neq \mathit{P}_j(y | x)$. This sort of scenario occurs when the input spaces perceived by the clients are analogous, but their outputs are not. A real situation, related to the training of an autonomous car, where this could happen is a yellow traffic light. When encountering a yellow traffic light, the correct output for some participants would be to stop the car, and for some others to continue driving without changes. This causes incompatibilities among the clients. 
    \item[iii)] $\mathit{P}_i(x) \neq \mathit{P}_j(x)$ and $\mathit{P}_i(y |x) \neq \mathit{P}_j(y | x)$. This is a combination of the two previous situations: Participants want to learn a common task, such as driving, but their input spaces are significantly unequal, and their reactions to some of the inputs are different too. 
\end{itemize} 

On the whole, this gives us a total of four different situations to account for. We represent them in Table~\ref{tab:tabla-espacial}, along with the different works that deal with each situation. There are lots of techniques that could fit the cell of IID data, such as FedAvg~\cite{mcmahan2016federated}. However, that situation is out of the scope of our work, and we will focus on the non-IID scenarios. Most of the works that consider heterogeneous data problems do not provide a classification of non-IID data, neither worry about the kind of heterogeneity they are trying to deal with. However, we locate them in Table~\ref{tab:tabla-espacial} according to the kind of non-IID data situation that they can solve. For this reason, we establish two possible classifications of the FL non-IID research.

The first way of classifying the different strategies is based on how the works place themselves in the FL context. Most of the FL works that deal with heterogeneous data describe their approach as a \emph{Personalization} approach to increase their accuracy over the different clients. This personalization can be performed at different levels. In Section~\ref{sec:personalization} we briefly explain these kinds of methods. It should be noticed that, although these strategies deal with data heterogeneity, they are unaware of which probability density function is varying in each situation.

On the other hand, once we have discussed and explained the different types of non-IID data that could exist, it seems very reasonable to classify the strategies according to the type of non-IID data it faces. This classification encompasses the other one, and at the same time it opens the door to consider other ML techniques that also deal with heterogeneous data and are close to FL. We describe these techniques in Section~\ref{sec:strategies-for-non-iid}. 

\begin{table}[ht]
    \centering
    \includegraphics[scale=0.45]{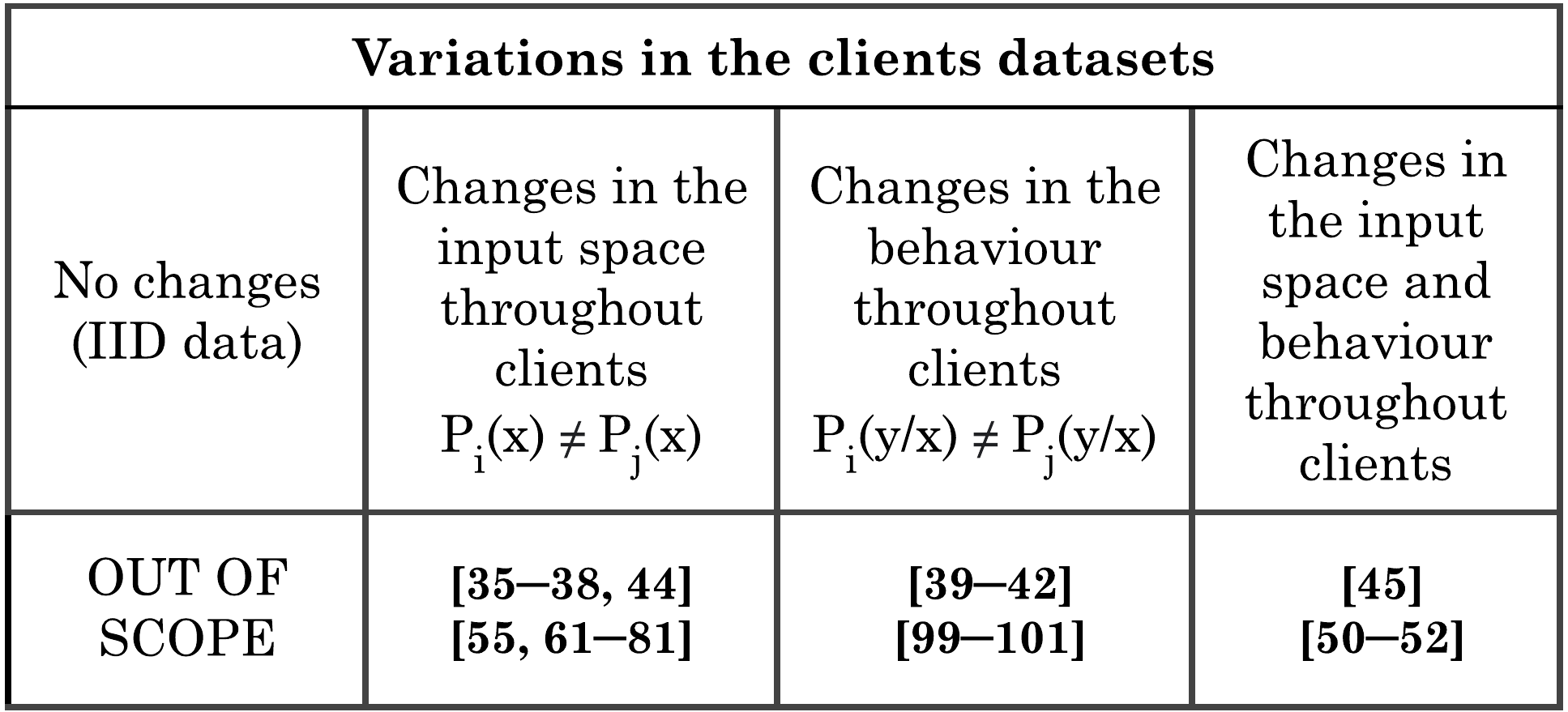}
    \caption{Non-IID learning scenarios in Federated Learning, and the strategies that could potentially solve each situation. Strategies that deal with changes in both the input space and the behaviour are placed only in the last column, and not in the previous ones.}
    \label{tab:tabla-espacial}
\end{table}

\subsection{State-of-the-art classification: Personalization strategies}
\label{sec:personalization}

In this section, we focus on the first classification we just mentioned. Some of the works present in the FL literature try to balance the generalized knowledge learned from the whole set of clients with the specificity of each of them. This kind of thinking gives rise to the personalization techniques, which precisely aim to grant more importance to the particular information of individual clients. It is empirically shown that in realistic situations one global model cannot fit the particularities of all clients~\cite{sattler2019robust}. In fact, some clients may have opposite interests sometimes, so we must open the door to the possibility of having more than just one global model. Personalization arises as an intermediate agreement between the model generalization and individualization, so that the model can learn not only the general knowledge but also the uncommon one.

Personalization can be implemented at different levels: each participant could have its own model, distinct from all of the others,  which we will refer to as \emph{Client-level Personalization}; or there could be groups of clients sharing the same model, i.e., \emph{Group-level Personalization}. Both options present some advantages, as they accomplish a better model performance, although their main drawback is that their computational requirements are more demanding.

\subsubsection{Client-level personalization}
\label{sec:clientlevelpersonal}

\emph{Client-level personalization} refers to the approaches that allow each participant to obtain its own model after the training process. When trying to personalize a global model trained in a distributed setting, a multitude of options and ideas have been studied and proposed. These approaches are gathered into two different classes. 

\begin{itemize}
    \item[i)] On the one hand, we find previously existent techniques of ML adapted to the FL framework to develop a better global model, such as Transfer Learning, or Multi-task Learning.
    \item[ii)] On the other hand, we encounter different implementations of a simultaneous two-level learning, local and global. Once the global model is obtained, each participant combines their local model, which was trained simultaneously with the global one, and attain their personalized model.
\end{itemize} 

The first type of solutions include proposals such as Transfer Learning techniques to adapt a pre-trained model over a public dataset to a bunch of devices~\cite{li2019fedmd}, although it can also be applied without needing a public dataset~\cite{wang2019federated}. Another idea, performed in~\cite{finn2017model,fallah2020personalized} is adapting algorithms from the Model-Agnostic Meta-Learning (MAML) setting, such as \emph{Reptile}, to a federated setting. In contrast with these approaches, some authors~\cite{yu2020learning,smith2017federated} consider the devices particularities constitute enough difference to assume that the training participants are performing different tasks, thus bringing up methods based on Multi-task Learning, such as MOCHA~\cite{smith2017federated}. All of the above ideas are reinterpretations of what one could understand by personalization, using different points of view to reuse already known techniques.

The latter class of approaches focuses on training two distinct models in parallel for each device. One of those two models is the global one, trained jointly by every client with its own data following the standard federated baseline, whereas the other one remains private and will be used to adjust the result from the federated training. The proposals vary, both in the way of achieving the global model and how the private model is taken into account. Here we present four alternatives. In~\cite{arivazhagan2019federated,nam2016learning}, clients train a Deep Neural Network (DNN), with the requirement that the last few layers are not shared, and each client trains them separately. In this case, the shared layers play the role of the global model, while these latter layers act as the personalization model, allowing different participants to obtain different results for similar inputs. In~\cite{hanzely2020federated}, clients follow probabilistic steps of training. A certain probability $p \in (0,1)$ is fixed at the beginning, and in each round of training participants execute locally one step of Stochastic Gradient Descent with probability $p$, and share the local models of their device to the server with probability $1-p$. Unlike in common federated frameworks, the global model is computed and distributed to each client, but they do not add the next updates over that model. Instead, they calculate a weighted mean using the global model and their local one, so in the end each client obtains a unique model.~\cite{li2018federated} focuses on training a model similar to FedAvg, but allowing some variation in the different local updates before they converge, and keeping a measure of the updates dissimilarities through the training process. At the end of the training procedure, each participant receives the global model, but it is allowed to modify it briefly according to their dissimilarity measure. This method is named \emph{FedProx}. To conclude, in~\cite{deng2020adaptive} devices train a global model as it is done in general in FL, but whenever a client participates in a training round, it trains a local model at the same time it trains the global one. Once training is finished, each client adjusts the global model obtained using the local model.

The experimental results of some of these proposals are quite remarkable. For instance,~\cite{finn2017model} compares their proposed meta-learning method with some other strategies of personalization, such as a fine-tuning baseline~\cite{tajbakhsh2016convolutional} and a k-nearest neighbour baseline~\cite{dudani1976distance}, obtaining an accuracy higher than these two algorithms by over 10\%.~\cite{arivazhagan2019federated} trains two well-known DNNs, \emph{MobileNet-v}1 and \emph{ResNet-}34, adding some personalization layers. They use the CIFAR-100 and FLICKR-AES~\cite{ren2017personalized} datasets, and split the data among clients imposing a non-iid division. They only compare both DNNs with standard FedAvg, however, they highly improve the accuracy, reaching 80\% in both datasets, while FedAvg barely gets 60\% with CIFAR-100, and 40\% with FLICKR-AES.~\cite{li2018federated} compares FedProx with FedAvg using the MNIST and Shakespeare datasets in both IID and non-IID situations. In the IID scenario, FedAvg performs slightly better. However, FedProx outperforms it by a huge difference in the non-IID setting, and also improves the convergence ratio in both cases. Lastly,~\cite{deng2020adaptive} uses the MNIST, EMNIST and CIFAR-10~\cite{krizhevsky2009learning} datasets to test the accuracy of their APFL~\cite{deng2020adaptive} strategy. They obtain an accuracy of 89\% over a non-IID split of the dataset, in opposition to FedAvg and FedAvg with fine-tuning, which obtains 32\% and 83\% of accuracy respectively.

\subsubsection{Group-level personalization}
\label{sec:grouplevelpersonal}

In contrast with attaining a distinct model for each client, another alternative is what we call \emph{Group-level personalization}, which consists of gathering the devices in clusters and training different models for each of them. This line of research emerged a couple of years ago, so it presumably has not achieved its full potential yet. For this reason, there are few different approaches to be mentioned, and all of them are very recent.

The most discussed topic in this area of research is how to split (or assemble) participants in order to get the groups that would benefit the most from sharing a model among them. One set of approaches are based on adopting hierarchical clustering techniques to partition the clients~\cite{sattler2020clustered,shlezinger2020communication,briggs2020federated}. Their strategy is based on using a measure of distance between the weight updates the clients send to the server to gather~\cite{sattler2020clustered,shlezinger2020communication} or divide~\cite{briggs2020federated} them. However, there is no mathematical guarantee that participants who send similar updates would benefit each other. It could be the case of some participants who collect very different data but the updates they generate are close to each other. Similarly, nothing ensures that the ones who send different updates would be better off apart. The only justifiable statement is that these participants would reach convergence faster than others, regardless of the performance of the attained model.

The other approach to perform this kind of personalization concerns the global data distribution, and the local distributions of the clients. The main point of the works concerning this line of research~\cite{Mohri2018,Mohri2019,Mohri2020three} is that if data is non-IID among the devices, then a shared global model cannot fit all of the data samples belonging to any client. When this happens, the global distribution may not represent the singularities of some participants, and thus the global model should not be trained according to that distribution. What these works propose is a method to determine the global distribution $\mathit{D}_\Lambda$ that best represents the different clients. This distribution does not necessarily coincide with the weighted mean global distribution (Equation~\ref{globaldistribution}). Once $\mathit{D}_\Lambda$ is found, clients are grouped according to some private parameters that depend on $\mathit{D}_\Lambda$. This strategy avoids the usage of the local updates to form the clusters and develops theoretical guarantees to justify that clustering the participants this way benefits the final model.

Concerning the experimental results,~\cite{sattler2020clustered,shlezinger2020communication,briggs2020federated} perform training on MNIST, FEMNIST, and CIFAR-100~\cite{krizhevsky2009learning} benchmark datasets, dividing the samples among the clients and swapping labels in some of them to produce different behaviours, $\mathit{P}(y|x)$. They achieve higher accuracy and convergence ratio than standard FedAvg. However, these methods do not compare themselves with any other algorithm than FedAvg, which is designed to tackle problems only in IID scenarios. On the other hand,~\cite{Mohri2018} improves the result obtained by FedAvg on the task of digit image recognition using the MNIST dataset, in a centralized framework.~\cite{Mohri2019,Mohri2020three} compare themselves with FedAvg in decentralized settings using the datasets of Fashion MNIST and Extended MNIST (EMNIST)~\cite{cohen2017emnist}, and they obtain a similar accuracy. On the whole, the most remarkable improvement accomplished with these kinds of personalization methods so far is their convergence speed. In Table~\ref{tab:datasets-personalization} we summarize the datasets used in these works.

\begin{table}[ht]
\centering
\resizebox{\columnwidth}{!}{%
\begin{tabular}{c|lll}
Article  & \multicolumn{3}{c}{Datasets used in experiments}                \\ \hline
\cite{li2019fedmd}                  & MNIST;                            & CIFAR-10       &             \\
\cite{finn2017model}                & MiniImageNet;                     & Omniglot       &             \\
\cite{fallah2020personalized}       & MNIST*;                           & CIFAR-10*      &             \\
\cite{yu2020learning}               & CASAS                             &                &             \\
\cite{arivazhagan2019federated}     & CIFAR-100;                        & FLICKR-AES \ \ &             \\
\cite{nam2016learning}              & OTB;                              & VOT2014        &             \\
\cite{li2018federated}              & MNIST;                            & FEMNIST;       & Shakespeare \\
\cite{deng2020adaptive}             & MNIST*;                           & CIFAR-10*;     & EMNIST      \\
\cite{sattler2020clustered}         & MNIST*;                           & CIFAR-100*     &             \\~\cite{shlezinger2020communication}  & MNIST*;                           & FEMNIST        &             \\
\cite{Mohri2019}                    & Fashion MNIST;  \ \               & EMNIST         &             \\
\cite{Mohri2020three}               & Fashion MNIST;                    & EMNIST         &             \\
\vphantom{a}
\end{tabular}
}
\caption{Summary of the datasets employed in the works presented in Section~\ref{sec:personalization}. Datasets marked with an asterisk are modified in different ways, making it impossible to fairly compare each other. Some of the datasets mentioned were not referenced so far: Omniglot~\cite{lake2011one}, OTB~\cite{wu2013online}, VOT2014~\cite{hadfield2014visual}, Shakespeare~\cite{mcmahan2017communication} and Fashion MNIST~\cite{xiao2017fashion}. }
\label{tab:datasets-personalization}
\end{table}

\subsection{Statistical taxonomy for non-IID strategies}
\label{sec:strategies-for-non-iid}

Once we have explained the existing personalization methods in FL, we now want to deepen into the other classification, based strictly on the kind of non-IID data the different works face. Going back to Table~\ref{tab:tabla-espacial}, we classify these methods into $2$ categories: those who work with changes in the input space throughout clients, i.e., changes in $\mathit{P}(x)$; and those who work with changes in the behaviour throughout clients, i.e., changes in $\mathit{P}(y|x)$. 

\subsubsection{Changes in the input space throughout clients}
\label{sec:spatial-inputspace}

\begin{figure*}[ht]
\begin{center}
\begin{forest}
for tree={
  minimum height=1cm,
  anchor=north,
  align=center,
  child anchor=north
},
[{Changes in the input\\space through clients},
  [{Domain\\transformation},
    [{Domains with\\particular features\\~\cite{zhao2019learning,liu2018multi,hoffman2012discovering}}]
    [{Domain\\factorization \\~\cite{siyahjani2015supervised,zhang2015feature,wang2010metric,weinberger2009distance}}]
  ]
  [{Personalization\\~\cite{li2019fedmd,wang2019federated,finn2017model,fallah2020personalized,li2018federated,Mohri2020three}},
  ]
  [{Domain adaptation\\~\cite{daume2009frustratingly,ganin2015unsupervised,you2019universal} },
    [{Dissimilarity\\methods\\~\cite{long2017deep,peng2019moment,pinheiro2018unsupervised,baktashmotlagh2013unsupervised}}]
    [{Sample\\reweighting\\ ~\cite{dredze2008online,dredze2010multi}}]
    [{Generative\\adversarial networks\\ \cite{tzeng2017adversarial,pei2018multi,cao2018partial,motiian2017few,peng2019federated}}]
  ]
]
\end{forest}
\end{center}
\caption{Classification of the different approaches that are able to solve the problem of spatial heterogeneity in the input spaces.}
\label{fig:taxonomy-spatial-input-space}
\end{figure*}
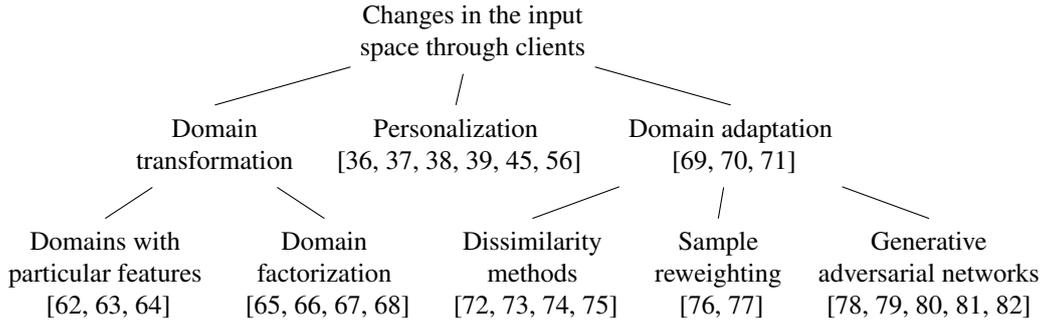

Each participant can collect data from their own input domain. However, in this Section we assume that their domains will remain unchanged during training, i.e., they will follow an IID data distribution over time.In decentralized settings, participants collect data independently, so nothing can assure their input domains are the same although that could be the case for some of them. This is by far the most studied kind of non-IID data, and strategies developed in this line of research pay attention to multiple problems that may occur in real-life scenarios. In this review, we include strategies that are not performed in the FL framework, but that are prone to be adapted to such settings. These works are classified into two main categories: \emph{(i) Domain Transformation} and \emph{(ii) Domain Adaptation}, which also branch into different approaches (see Fig.~\ref{fig:taxonomy-spatial-input-space}). 

Concerning the first of them, Domain Transformation methods focus on detecting the particularities and the common parts of the data domains, and they try to build a new shared input space. After that, the input spaces perceived are transformed into a common input space. As far as we are concerned, this strategy has only been performed in centralized settings. Nonetheless, in a decentralized setting each participant could carry out these calculations and the central server could build the common input space. 

A serious difficulty when trying to deploy this kind of method is that feature spaces in realistic problems tend to be very high-dimensional spaces, and that causes problems such as needing more processing capacity, or inaccurate results due to the curse of dimensionality. For instance,~\cite{zhao2019learning,liu2018multi,hoffman2012discovering} consider each domain may have its own set of features to characterize the samples, causing incompatibilities across domains, and they develop methods to extract a common feature representation. A different approach, performed in~\cite{siyahjani2015supervised,zhang2015feature,wang2010metric,weinberger2009distance}, consists of  constructing a factorization of the feature space with some properties.~\cite{siyahjani2015supervised,zhang2015feature} split the feature space into two orthogonal subspaces: one of them contains the domain variations, whereas the other one keeps the common parts, and both are used separately to perform learning.~\cite{wang2010metric,weinberger2009distance} divide the input space into an arbitrary number of low-dimensional spaces and apply techniques of Distance Metric Learning (DML)~\cite{yang2006distance,xing2002distance} in each of them.

On the other hand, Domain Adaptation methods~\cite{daume2009frustratingly,ganin2015unsupervised,you2019universal} work with a distinct situation, close to Transfer Learning, and they are in general deployed in centralized frameworks. However, some works aligned with this line of research talk about Federated Transfer Learning~\cite{chen2020fedhealth}, and consider the FL settings. In these cases, users are assumed to train a model with their data belonging to some domains, but they then apply that same model to get predictions on other domains. In this kind of setting, it is commonly said that the samples drawn for training belong to \emph{Source Domains}, while the samples used for prediction belong to \emph{Target Domains}. The most important difference with the previous category is that in this case there are no training data from the Target Domains, so it is not possible to create an appropriate feature space for learning those domains. Some other works in this line of research~\cite{long2017deep,peng2019moment,pinheiro2018unsupervised,baktashmotlagh2013unsupervised} present a variety of methods to measure the dissimilarity between the source and target domains, such as Maximum Mean Discrepancy (MMD)~\cite{long2017deep}, or Moment Matching for Multi-source DA (M$^3$SDA)~\cite{peng2019moment} and adjust the training model in an unsupervised manner \cite{pinheiro2018unsupervised,baktashmotlagh2013unsupervised}.

Another possibility for adapting the domains that also consists on calculating distances among the data distributions, make use of that information for re-weighting the samples from the closest ones to improve the model performance.~\cite{dredze2008online,dredze2010multi} apply this method using the Kullback-Leibler~\cite{van2014renyi} divergence, a well-known metric from information theory. 

A different approach for Domain Adaptation is based on Generative Adversarial Networks (GANs)~\cite{tzeng2017adversarial,pei2018multi,cao2018partial,motiian2017few,peng2019federated}. This strategy, that achieves remarkable results, consists of training two neural networks simultaneously: one of them is designed to create fake input data from the different domains, and the other one aims to distinguish the real data samples from the fake ones. Article \cite{peng2019federated} is particularly interesting because they employ GANs in federated settings. In these works, the method presented is compared with some other pre-existing methods, such as  Deep Adaptation Network (DAN)~\cite{long2015learning}, Deep Domain Confusion (DDC)~\cite{tzeng2014deep} and Residual Transfer Network (RTN)~\cite{long2016unsupervised}. These are some state-of-the-art techniques in Domain Adaptation. However, GANs methods outperform them in well-known tasks, such as object recognition using the Office-31~\cite{saenko2010adapting}, Office-Home~\cite{venkateswara2017deep} and VisDA2017~\cite{peng2017visda} datasets, and digit recognition using the MNIST~\cite{lecun1998gradient}, USPS and SVHN~\cite{netzer2011reading} datasets. 

In general, comparing the different methods is a tough issue. Each work is free to choose different synthetic datasets to perform experimental results, and also modify them to generate the required heterogeneity that they want to face (see Table~\ref{tab:datasets-inputspacethroughclients}). For these reasons, it is impossible to fairly compare the diverse strategies we presented. However, there are some remarkable results that we would like to highlight: regarding Domain Transformation methods, the experimental results of two of the works stand out~\cite{hoffman2012discovering,weinberger2009distance}. They present a complete variety of experiments and contrast their results with other well-known methods, getting significantly better error ratios and accuracies. On the other hand, the most outstanding results achieved with Domain Adaptation methods are the ones from~\cite{pinheiro2018unsupervised,cao2018partial,motiian2017few}. The first one,~\cite{pinheiro2018unsupervised} propose their method SimNet and experimentally compare their results with some other methods like DAN, RTN and a baseline method over the datasets of MNIST, Office-31 and VisDA2017. It improves the accuracy obtained by every other method in the three cases. Concerning~\cite{cao2018partial,motiian2017few}, they both employ the Office-31 dataset, and obtain impressive results compared to the other methods they test.

Besides all of the strategies we just talked about, there are a bunch of other methods to deal with the domain shifts. One of them is~\cite{gopalan2011domain}, which also mentions the Source and Target Domains, but also factorizes the input space to search for a Grassmann Manifold that fits all of the data samples. Afterward, the training is performed only on that manifold, instead of in the whole feature space. Lastly, some of the federated strategies of personalization explained in Section~\ref{sec:personalization} can also deal with the kind of heterogeneity brought up in this Section~\cite{li2019fedmd,wang2019federated,finn2017model,fallah2020personalized,li2018federated,Mohri2020three}.

\begin{table}[ht]
\centering
\resizebox{\columnwidth}{!}{%
\begin{tabular}{c|lll}
Article  & \multicolumn{3}{c}{Datasets used in experiments}                \\ \hline    
\cite{zhao2019learning}             & MNIST + SVHN + USPS              &                 &              \\
\cite{hoffman2012discovering}       & Office-31;                       & Bing-caltech256 &              \\
\cite{wang2010metric}               & COREL5000;                       & Trecvid2005\footnotemark[1]   &\\
\cite{weinberger2009distance}       & MNIST*;                          & Olivetti FR\footnotemark[2]   &\\
\cite{ganin2015unsupervised}        & MNIST + SVHN                     &                 &              \\
\cite{you2019universal}             & Office-31;                       & ImageNet;       & VisDA2017    \\
\cite{long2017deep}                 & Office-31;                       & Image CLEF-DA   &              \\
\cite{peng2019moment}               & Digit5;                          & Office-31       &              \\
\cite{pinheiro2018unsupervised}     & MNIST + MNIST-M + USPS; \ \      & VisDA2017       &              \\
\cite{tzeng2017adversarial}         & MNIST + SVHN + USPS;             & Office-31;      & NYUD         \\
\cite{pei2018multi}                 & Office-31;                       & Image CLEF-DA\footnotemark[3] &\\
\cite{cao2018partial}               & Office-Home;                     & VisDA2017             &        \\
\cite{motiian2017few}               & MNIST + SVHN + USPS;             & Office-31             &        \\
\vphantom{a}                                                                                
\end{tabular}
}

\caption{ Summary of the Datasets employed in the works presented in Section~\ref{sec:spatial-inputspace}. Asterisks indicate that the datasets have been modified in particular ways, making it impossible to fairly compare each other. Some of the datasets mentioned were not referenced so far: Bing-caltech256~\cite{bergamo2010exploiting}, COREL5000~\cite{chen2006miles}, ImageNet~\cite{russakovsky2015imagenet} and NYUD~\cite{silberman2012indoor}.  }
\label{tab:datasets-inputspacethroughclients}
\end{table}

\subsubsection{Changes in the behaviour throughout clients}
\label{sec:spatial-behaviours}

Differences in the behaviour of the clients refer to discrepancies in their conditional probabilities $\mathit{P}(y | x)$. A variation of this nature means that for, at least for some data samples, the correct output is not the same for all of the clients. More formally:

\begin{definition}
$2$ clients $i, j \in \mathit{N}$ present different conditional probabilities $\mathit{P}(y | x)$ if there exist a significant quantity $T$ of data samples $\{ x_k \}_{k=1}^T$, and distinct outputs $\{ y_{k_1}, y_{k_2} \}_{k=1}^T$ such that the participants $i, j $ own the data samples $(x_k,y_{k_1})$ and $(x_k,y_{k_2})$ respectively, $\forall \ k \in \{1,\ldots, T\}$.
\end{definition}

\begin{figure}[ht]
    \centering
    \includegraphics[scale=0.9]{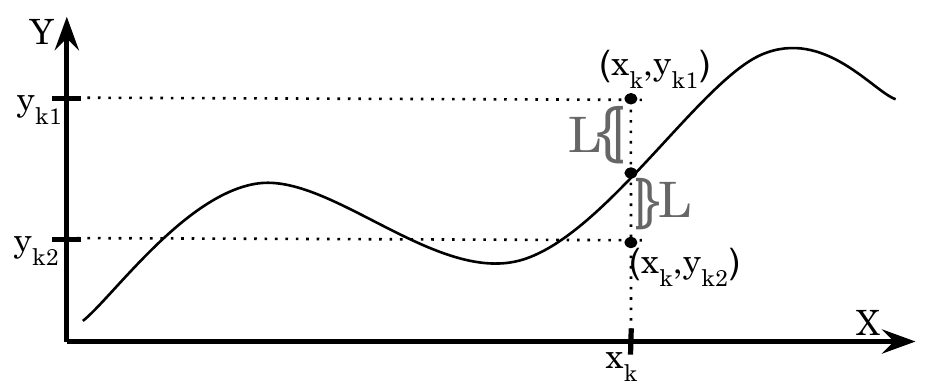}
    \caption{Regression model in one variable. The samples $(x_k,y_{k_1})$ and $(x_k,y_{k_2})$ belong each to one different client. If the distance between those samples is bigger than $2L$, there is no output that could be closer than $L$ for both of them.}
    \label{fig:different-behaviours}
\end{figure}

To be precise we need to specify the meaning of ``distinct outputs": $y_{k_1}, y_{k_2}$ are distinct if $ || y_{k_1} - y_{k_2} || > 2L$ for a certain margin of error $L$ (see Fig.~\ref{fig:different-behaviours}), which may vary depending on the specific problem. In a classification problem $2L$ needs to be lower than $1$, so the margin of error has to be less than $1 / 2$, whereas in regression problems the margin of error is fixed depending on the level of accuracy desired.

\footnotetext[1]{Available at http://www-nlpir.nist.gov/projects/trecvid}
\footnotetext[2]{Available at http://www.uk.research.att.com/facedatabase.html.} 
\footnotetext[3]{Available at http://imageclef.org/2014/adaptation}

The main problem regarding this context is that, unlike the previous one, a single global model cannot fit all of the users behaviours, since it will not be able to produce different predictions for the same input data. A model in ML is, by definition, a mapping $\mathit{M} : \mathit{X} \longrightarrow \mathit{Y}$ that assigns one value to each possible input~\cite{zhang2020machine}. Given the input $x_k$, a traditional model in a distributed setting would process it equally for all clients, thus predicting one, and only one, output. As a result, the predicted value could be $y_{k_1}$, $y_{k_2}$ or a different one, possibly intermediate, but in any case one of the clients would obtain a prediction with an error bigger than $L$. In addition, to detect the existence of several behaviours in the clients, we need to study the result of their loss functions, which implies having a certain amount of labeled data.

\begin{figure}[ht]
\begin{center}
\resizebox{\columnwidth}{!}{%
\begin{forest}
for tree={
  minimum height=1cm,
  anchor=north,
  align=center,
  child anchor=north
},
[{Changes in the behaviour\\ through clients},
  [{Group-level\\personalization\\methods\\~\cite{yu2020learning,arivazhagan2019federated,nam2016learning,deng2020adaptive,briggs2020federated}}]
  [{Contextual\\Information\\methods\\~\cite{yang2014unified}}]
  [{Federated\\Multi-Task\\Learning\\~\cite{smith2017federated,corinzia2019variational}}]
  [{Cohort-based\\Federated\\Learning\\~\cite{hiessl2021cohort}}]
  %[{Error-based\\methods}]
]
\end{forest}
}
\end{center}

\caption{Classification of the different approaches that deal with the spatial heterogeneity in the behaviours of clients.}
\label{fig:taxonomy-spatial-behaviour}
\end{figure}
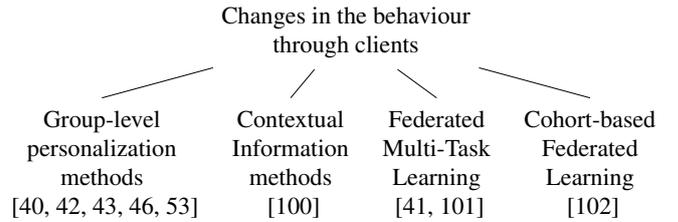

To overcome this matter, it is essential to consider some kind of model architecture that grants the possibility of variations among the participants. That is, making it possible for each user $i \in \mathit{N}$ to have a somehow personalized model $\mathit{M}_i$, distinct from the others. Some of the strategies of personalization discussed in Section~\ref{sec:federated-non-iid} can be adapted to deal with different behaviours (see Fig.~\ref{fig:taxonomy-spatial-behaviour}). For instance, having a proper metric of the error would allow the \emph{Group-level personalization} strategies to cluster the participants according to their behaviour. This approach is very similar to Cohort-based Federated Learning~\cite{hiessl2021cohort}, which precisely organize clients into cohorts with very similar data distributions. 

There are few approaches specifically designed to deal with this kind of non-IID data. One of the closest techniques to tackle this issue, although it does not specifically talk about FL, is the one deployed in~\cite{yang2014unified}. This strategy consists of including additional pieces of information to the input data, such as a task identifier $z$. This allows $2$ very similar data samples $x_{k_1}, x_{k_2}$ to be distinct: $(x_{k_1},z_1), (x_{k_2},z_2)$.

To conclude, there are also works in the crossroad between FL and Multi-task Learning~\cite{smith2017federated,corinzia2019variational}. The first one was already discussed in Section~\ref{sec:personalization}. The latter also presents the multi-task framework combined with the federated setting. It performs experiments using the MNIST, EMNIST, and Shakespeare datasets, and compares itself with well-known federated algorithms such as FedAvg and FedProx.

\section{Non-IID Data in Continual Learning: Concept Drift}
\label{sec:continual}

In this section, we are going to consider \emph{Continual Learning (CL)} problems, which involve the training of models over time. In the standard ML setting, the objective is to build a prediction model using a certain amount of data. A key point to discuss is that the training dataset is typically assumed to be fully available from the beginning, and this may conflict with realistic situations, where data is collected progressively and changes over time. For that reason, it is convenient to talk about CL, a ML setting in which models continuously learn and evolve using new streams of data samples, while aiming to retain preceding concepts. This kind of framework has been given different names over the years~\cite{lesort2020continual}, like \emph{Lifelong Learning}~\cite{thrun1995lifelong,tessler2017deep}, \emph{Never Ending Learning}~\cite{carlson2010toward,mitchell2018never} and \emph{Incremental Learning}~\cite{xiao2014error,gepperth2016incremental,rebuffi2017icarl}, but all of them rely on the same ideas: training a model gradually with data collected over different periods of time, adapting to the new instances and trying to preserve the previous knowledge. 

We introduce the CL framework because we aim to talk about the time-evolving condition of FL problems. However, throughout this section we are going to cite and briefly describe works focused on CL that do not necessarily consider the FL framework. This is because, as we already mentioned, there are almost no works that focus on both FL and CL simultaneously~\cite{usmanova2021distillation,park2021tackling,yoon2021federated}. Nonetheless, the works we consider are, from our point of view, the ones that would be more easily adaptable to the FL framework. 

Training a model using CL techniques presents some specific problems, which have already been studied in recent literature. The most challenging ones are, as it occurred with FL, related to the data distribution. CL was conceived as a centralized paradigm of ML so, even though non-IID data across devices has not been discussed nor handled so far, it can evolve in time. This is a complication, as the model could be unable to converge to a solution if the training data shifts constantly. Another undesirable situation, named \emph{catastrophic forgetting}, is that the model completely and abruptly forgets previously learned concepts if they are not present in the current data anymore~\cite{kemker2018measuring,goodfellow2013empirical}. For these reasons we are going to focus on how data behaves as time goes by, and how to act if the data shifts drastically, in unpredictable ways. This is commonly known as \emph{concept drift}~\cite{gama2014survey,lesort2020continual}.

\subsection{Concept drift definition}
\label{sec:drift-definition}

The non-stationary distribution is caused by changes in data over time. These changes can be seen as variations in the frequencies certain kind of data appears: a concept has frequency zero if it has not appeared yet in the dataset, and when it shows up its frequency becomes a positive number. This kind of variation, called \emph{concept drift}, is one of the most important CL challenges~\cite{lesort2020continual,khamassi2018discussion}. We can formally define them as follows: 

\begin{definition}
Given a time period $[0,t]$, and a set of samples $S^{0,t} = \{ (x_j,y_j) \}_{j=0}^t$ with a certain probability distribution $D^t(x,y)$, where $x_j$ is a feature vector and $y_j$ is its correspondent output; we say a Concept Drift occurs at timestamp $t$ if there is a significant difference between $D^t(x,y)$ and $D^{t+1}(x,y)$: 
\[ \exists \ t: D^t(x,y) \nsim D^{t+1}(x,y). \] 
\end{definition}

Note that, if the problem we are trying to deal with is by nature a federated problem that evolves in time, each client might experiment a drift in different moments. Also, one local concept drift does not necessarily have an impact on the global distribution. It may be the case that a local drift on client $i$ results in a change in the distribution of $i$, but not in the joint distribution, $D_G^t(x,y)$. This is an example of why concept drifts are potentially dangerous for the model performance, and hence must be detected and counteracted.

\begin{figure*}[ht]
    \centering
    \includegraphics[width=0.65\textwidth]{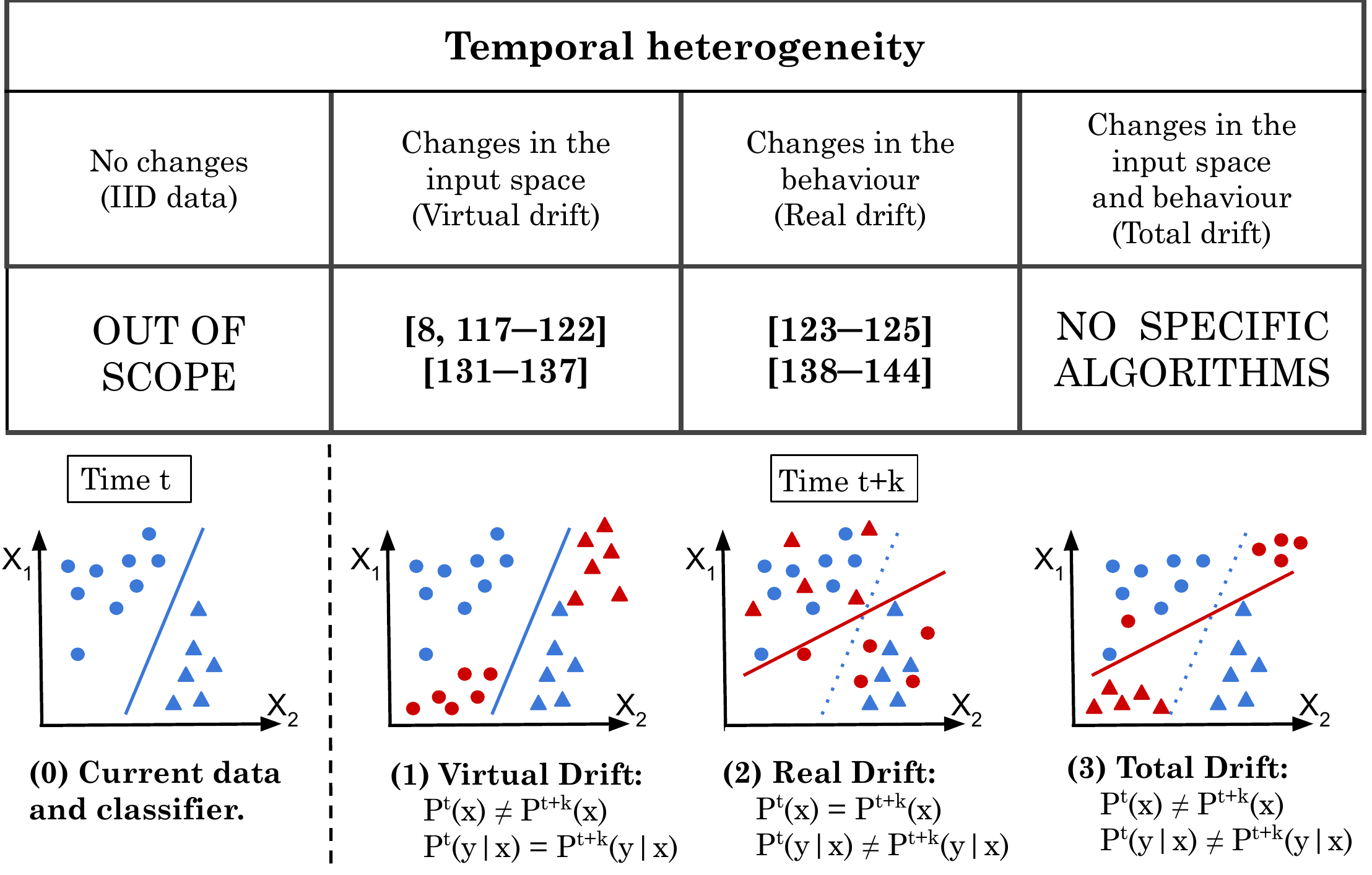}
    \caption{Representation of Concept drifts in a two-dimensional input space $X$ with two possible labels $Y = \{ \circ, \vartriangle \}$. On the left of the doted line, we find the data samples received before time $\mathbf{t}$, and on the right there are three possible time-evolving situations. (1): the new data samples observed are situated in new regions, previously unseen. However, data labels correspond with the split made by the classifier from (0). (2): the new instances appear in already known regions of the input space, but they are incorrectly classified using the model from (0). (3): the two previous situations are combined.}
    \label{fig:virtual_and_real_drifts}
\end{figure*}

When trying to deal with concept drifts, one should notice that not all of them are alike, as data can evolve in multiple ways. Similar to what occurred with non-IID data in FL, it is important to characterize concept drifts to distinguish them. In the case of concept drifts, most of the existing works present a common ground, and base their classification according to which factor from the equation $\mathit{P}(x,y) = \mathit{P}(x) \cdot \mathit{P}(y | x)$ is altered. According to this criteria, we determine three types of shift~\cite{gepperth2016incremental,webb2016characterizing}: (1)~virtual, (2)~real, and (3)~total (see Fig.~\ref{fig:virtual_and_real_drifts}):

\begin{itemize}
    \item[i)] \emph{Virtual concept drift} makes reference to variations in just the marginal probability density, $\mathit{P}^t(x) \neq \mathit{P}^{t+k}(x)$ and $\mathit{P}^t(y|x) = \mathit{P}^{t+k}(y|x)$. Returning to the example used in Section~\ref{sec:non-iid-classification} of training an autonomous car, this situation happens, for instance, when clients move into places or regions previously unseen for them.
    \item[ii)] \emph{Real concept drift}, related to differences in conditional probabilities, $\mathit{P}^t(x) = \mathit{P}^{t+k}(x)$ and $\mathit{P}^t(y|x) \neq \mathit{P}^{t+k}(y|x)$, is caused by a change in the conditional probability of the classes with respect to the input features, i.e., similar input data samples that have unequal labels. An example of this would be, again, a yellow traffic light. Sometimes a client would stop the car when encountering a yellow traffic light, and some others may continue driving. 
    \item[iii)] \emph{Total concept drift} is the mixture of the two other drifts, $\mathit{P}^t(x) = \mathit{P}^{t+k}(x)$ and $\mathit{P}^t(y|x) = \mathit{P}^{t+k}(y|x)$, and it is the result of both probabilities evolving significantly over time.
\end{itemize}   

This classification is analogous to the one we proposed in Section~\ref{sec:federated-non-iid} for the data heterogeneity across clients. In the scenario we consider, the task remains unchanged, and therefore it is a single-incremental-task scenario~\cite{lesort2020continual}. Apart from these cases we just discussed, concept drift also takes place when the task itself changes, since that also modifies $D^t(x,y)$. This scenario is closely related to multi-task learning~\cite{smith2017federated,zhang2021survey,caruana1997multitask}.

\subsection{Concept drift detection}
\label{sec:drift-detection}

Once we settled what we understand by concept drift, we can discuss the methods developed to deal with it. Those methods typically consist of three parts. The first step is that they need to detect modifications in the distributions. Then, they have to act in consequence to the detected changes, so the model obtained is adjusted to the current scenario. Finally, it is important to explain the drift and understand their implications for future training. In this Section we just examine the detection strategies. The algorithms that implement a response to these drifts will be reviewed in Section~\ref{sec:adressing-non-iid-data}. 

Many concept drift detection strategies have been proposed to attach the situations of virtual and real drifts~\cite{casado2021concept,hammoodi2018real,dries2009adaptive,shao2014prototype,gama2004learning,baena2006early,manias2021concept,bifet2009adaptive,bifet2009improving,frias2014online}. These approaches are often classified as \emph{Data Distribution-based} or \emph{Error Rate-based} methods correspondingly. They use different statistical properties of the input and output distributions to identify their breaking points, corresponding to drifts. Most of these strategies of concept drift detection consider a situation where data is centralized in one single machine. As far as we are concerned, the only works that present a concept drift detection strategy on a federated setting are~\cite{casado2021concept,manias2021concept}. Nonetheless, the strategies we highlight are, from our point of view, easily adaptable to the FL framework.

When trying to detect a virtual concept drift, the only required information is the input pattern of the data samples, $\{x_i \}_{i=1}^M$, or some transformation of it. For instance, the strategy developed in~\cite{hammoodi2018real} works directly with the input data, and is based on measuring the similarities among the features, grouping them in clusters, and evaluating the number of features from the new data sample in each cluster to identify a drift. On the other hand,~\cite{dries2009adaptive} works with an alteration of the input data. They determine a mapping that relies on the input features of the samples $f: \mathit{X} \longrightarrow \{-1,1\}$ and apply it to the whole input dataset, splitting it into two groups (the ones that go to $1$ and the ones that go to $-1$). Then, they statistically compare if data received before and after a certain timestamp is equally distributed in those groups. If they are not, a drift is detected. Another recurrent strategy for the detection of virtual drifts consists of using sliding windows to keep track of the samples received in the past and compare them to the current data stream~\cite{casado2021concept,shao2014prototype}.

In contrast with virtual concept drift detection, real ones present more of a challenge. In the first place, the virtual concept drift strategies we just presented can be deployed in unsupervised settings since they do not need any label information, whereas real drift detection methods need it because the main variable involved when trying to detect changes in conditional probabilities, $\mathit{P}(y | x)$, is the error in the predictions. Some of these works also employ sliding windows to perform the drift detection~\cite{bifet2009adaptive,bifet2009improving}, although in general techniques aiming to detect this kind of variations are highly dependant on how the model inaccuracy is measured~\cite{frias2014online}. There are different ways of accounting for the model loss. One of the most extended functions for measuring the error in machine learning models is the well-known cross-entropy loss. A lot of research has been made to determine whether this is an appropriate measure of the conducted error~\cite{nar2019cross,feng2020can}. In addition, different authors have proposed many other loss functions based on the cross-entropy loss~\cite{ho2019real,zhang2018generalized,li2019dual}. All of these alternatives present some limitations, such as weaknesses against skew labeled data, or the fact that errors are untraceable. These kinds of properties are very desirable when facing real drift, as they provide important information about the origin of the error.

\section{Addressing Federated and Continual non-IID data}
\label{sec:adressing-non-iid-data}

For what we have seen in Section~\ref{sec:continual}, concept drift in CL scenarios can be interpreted as the counterpart of \emph{non-IID data} in the FL ones, i.e., changes in the distribution as time passes are the origin of statistical heterogeneity in continual settings. Notice that variations on the distribution of one dataset over time should be always contemplated as identically distributed, since there is only one dataset affected. Nonetheless, the sets of data collected by one client $i$ over time, corresponding to different timestamps, $\mathit{D}_i^t, \mathit{D}_i^{t+k}$, could be studied as two different datasets, $ \mathit{D}^t \subsetneq \mathit{D}^{t+k}$. 

\begin{table*}[ht]
    \centering
    \includegraphics[width=0.7\textwidth]{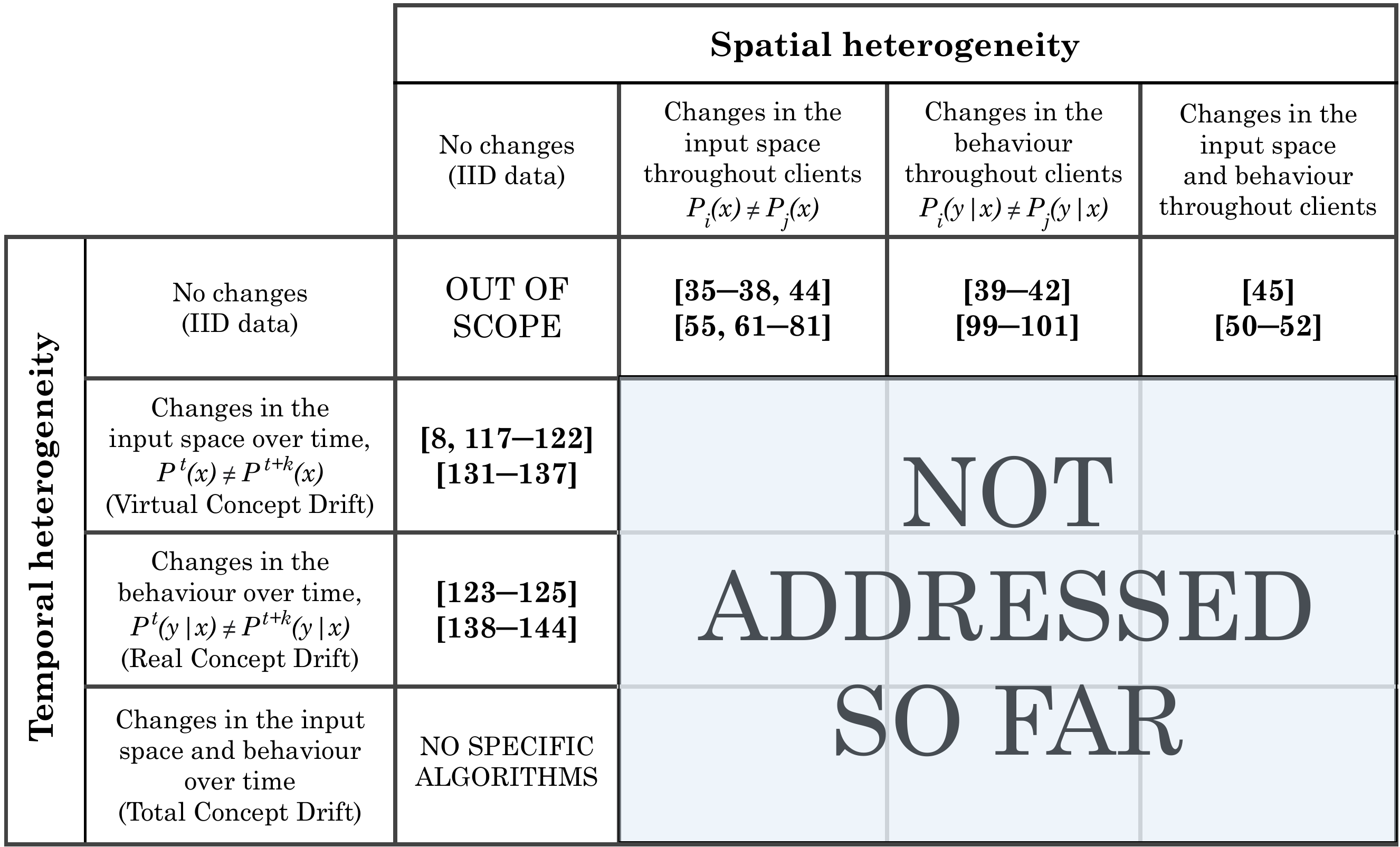}
    \caption{Spatial and Temporal heterogeneity learning scenarios, and the strategies that could potentially solve each situation. Strategies that deal with changes in both the input space and the behaviour are placed only in the last row/column, and not in the previous ones.}
    \label{tab:tabla-cambios}
\end{table*}

In that sense, it is logical to talk about non-identical distributed data over time. In fact, considering again the factorization $\mathit{P}(x,y) = \mathit{P}(x) \cdot \mathit{P}(y | x)$, the casuistry is identical to the one explained in Section~\ref{sec:federated-non-iid}: if the client data distribution is stationary (IID over time), then we can ensure that both factors remain equal over time: $\mathit{P}_i^t(x) = \mathit{P}_i^{t+k}(x)$ and $\mathit{P}_i^t(y | x) = \mathit{P}_i^{t+k}(y | x)$. Else, we find the three possibilities illustrated in Fig.~\ref{fig:virtual_and_real_drifts}. From now on, we will call \emph{temporal non-IID data} to the data heterogeneity that a client can undergo over time (see Section~\ref{sec:drift-definition}). Conversely, we will call \emph{spatial non-IID data} to the data heterogeneity across clients that are training a shared model (see Section~\ref{sec:federated-non-iid}).

In real-life problems, data distributions can vary in a bunch of different ways. Clients in a federated setting are expected to collect their own data samples, under particular conditions, leading to statistically unequal datasets. These differences can rely on the inputs each client perceive, $\mathit{P}_i(x) \neq \mathit{P}_j(x)$, as well as on the label associated with their inputs, $\mathit{P}_i(y|x) \neq \mathit{P}_j(y|x)$ (see Table~\ref{tab:tabla-espacial}). If we desire the model to be adapted to the particularities of the training participants, standard FL techniques will not be enough. Moreover, the process of collecting data and solving a task takes a certain amount of time, so the desired model should be able to evolve and adjust to future situations. Data will be collected during a long period of time, leading to changes in the input space, $\mathit{P}^t(x) \neq \mathit{P}^{t+k}(x)$ and also in the labels, $\mathit{P}^t(y|x) \neq \mathit{P}^{t+k}(y|x)$ (see Fig.~\ref{fig:virtual_and_real_drifts}).

On the whole, there are $4$ feasible scenarios for each spatial and temporal data, and they may appear combined with each other in realistic tasks. The global data distribution $D_G^t(x,y)$, which includes both spatial and temporal heterogeneity, can evolve following $16$ different courses. We represent all of the possibilities in Table~\ref{tab:tabla-cambios}, as well as some of the strategies and algorithms that focus on solving some of those possibilities. Notice that we include IID data to consider all of the possible combinations of heterogeneity.

It is reasonable to think that each course must be faced with specific methods. For instance, if the problem we are considering presents changes in the input space over time across one or multiple participants, it could be solved using a memory-based method to generalize the data from previous distributions and avoid catastrophic forgetting. Despite fitting perfectly for this problem, these kinds of solutions cannot deal with changes in behaviours over time, or changes in the input space across clients. For this reason, in this Section we focus on determining which strategies are more suitable to deal with each circumstance. For now, we only explained some algorithms from FL that were proposed as \emph{Personalization strategies}, without further explanation about the origin of the data heterogeneity; as well as methods to detect drifts, but not to react to them. 

Notice that if we determine effective approaches to solve each of the scenarios corresponding to the first row and column in Table~\ref{tab:tabla-cambios}, then we will be able to solve the situation of any cell by combining the algorithms from its corresponding row (already addressed in Section~\ref{sec:spatial-inputspace}) and column, as long as they are compatible. Thus, from now on we will consider scenarios where data is IID in the spatial axis. This corresponds to pure CL. In the following sections, we are going to present the existing solutions to deal with temporal non-IID data, classify those strategies according to their shared characteristics, and compare their experimental results when possible. 

As far as we are concerned, there are only few methods for addressing both federated and continual issues at the same time~\cite{usmanova2021distillation,park2021tackling}. However, there is still room for significant contributions in these types of scenarios and a lot of situations that have not been taken into account. 

\subsection{Virtual Concept Drifts} 
\label{sec:temporal-non-iid-input}

\begin{figure}[ht]
\begin{center}
\resizebox{\columnwidth}{!}{%
\begin{forest}
for tree={
  minimum height=1cm,
  anchor=north,
  align=center,
  child anchor=north
},
[{Virtual concept drift},
  [{Memory-based\\methods}, 
    [{Rehearsal methods\\~\cite{rolnick2018experience,chaudhry2019continual}}]
    [{Generative methods\\~\cite{shin2017continual,van2018generative}}]
    ]
  [{Regularization\\methods\\~\cite{kirkpatrick2017overcoming,schwarz2018progress,ritter2018online}}]
]
\end{forest}
}
\end{center}

\caption{Classification of the different techniques able to deal with the temporal heterogeneity in the input space of data.}
\label{fig:taxonomy-temporal-input-space}
\end{figure}
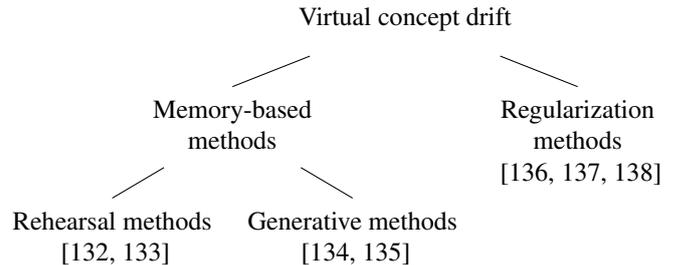

As we already discussed in Section~\ref{sec:drift-detection}, the procedures to detect virtual drifts rely only on the input data distributions. Similarly, the approaches to prevent the model from lowering its accuracy involve just the data samples. They can be classified in \emph{Memory-based methods} and \emph{Regularization methods}, see Fig.~\ref{fig:taxonomy-temporal-input-space}.

 Memory-based methods~\cite{rolnick2018experience,chaudhry2019continual,shin2017continual,van2018generative} focus on keeping a record of data samples from the previous concepts, so when a drift is detected, the network is trained both with new data and the data recorded to avoid forgetting.~\cite{rolnick2018experience} proposes a method (CLEAR) to store the data samples already used for training and use them again in the future, combined with the new collected data. They show that this strategy is quite effective to prevent catastrophic forgetting.~\cite{chaudhry2019continual} proposes a similar technique, ER, but with the difference that they just store little amounts of data, and use them repeatedly after drifts mixed with the new samples. They conclude that their approach neither harm generalization nor causes overfitting to the saved data samples. 

There are other existing alternatives encased in the Memory-based methods which consist of using the recorded data to generate similar samples~\cite{shin2017continual,van2018generative}. This kind of technique tries to avoid the possibility of model overfitting to the specific samples stored in memory.~\cite{shin2017continual} implements a Generative Adversarial Network (GAN) to sample the new instances of data when drifts are detected. However, training both the main model and the GAN could result computationally expensive. One way of addressing this situation without needing a second network is integrating the generative model into the main model by equipping it with generative feedback connections~\cite{van2018generative}.

On the other hand, \emph{Regularization methods}~\cite{kirkpatrick2017overcoming,schwarz2018progress,ritter2018online} impose restrictions in the weight updates to keep them fixed and avoid forgetting an input domain that has already been learnt.~\cite{kirkpatrick2017overcoming} propose a state-of-the-art method denominated Elastic Weight Consolidation (EWC), which learns a task, and then determines which connections between weights are determinant to correctly perform that task. Those connections are set applying the matrix of Fisher Information, a statistical tool that weighs the relevance and contribution of each weight to the final result of the model. To do so, it estimates the probability $\mathit{P}(x | y)$, i.e., determines the distribution of inputs as a function of their classes. This approach is usually seen in other works as a baseline method for avoiding catastrophic forgetting. However, it presents some issues, such as scalability and computational efficiency when trying to learn several tasks.~\cite{schwarz2018progress} introduces a new technique based on EWC which partially solves those issues. Similarly,~\cite{ritter2018online} also consider the conditional probability $\mathit{P}(x | y)$, but in this case they use a Laplacian approximation to reduce the computational costs involved.

\begin{table}[ht]
\centering
\resizebox{\columnwidth}{!}{%
\begin{tabular}{c|ll}
Article  & \multicolumn{2}{c}{Datasets used in experiments}                \\ \hline
\cite{chaudhry2019continual}        & MNIST* + CIFAR-10*;             & CUBS        \\
\cite{shin2017continual}            & MNIST + SVHN                    &             \\
\cite{van2018generative}            & MNIST*                          &             \\
\cite{kirkpatrick2017overcoming}    & MNIST*                          &             \\
\cite{ritter2018online}             & MNIST* + SVHN + CIFAR-10  \ \ \ &             \\
\vphantom{a}
\end{tabular}
}
\caption{ Summary of the datasets employed in the works presented in Section~\ref{sec:temporal-non-iid-input}. Asterisks indicate that the datasets have been modified in particular ways, making it impossible to fairly compare each other.}
\label{tab:datasets-inputspaceovertime}
\end{table}

Concerning the experimental results, we find the same problem of having very different datasets (see Table~\ref{tab:datasets-inputspaceovertime}), making it difficult to establish relations between the different strategies results. In this case, nonetheless, most of the works we just presented compare their methods with EWC~\cite{kirkpatrick2017overcoming}, using it as a baseline. In~\cite{kirkpatrick2017overcoming}, the MNIST dataset is employed to simulate the different input data distributions. They take a random permutation of pixels and apply that permutation to all of the images to create each input domain. With this strategy, they only need one dataset, denoted Permuted MNIST, to simulate any number of domains. In~\cite{schwarz2018progress}, authors compare the accuracy obtained with their learning method (P\&C) in each domain, and also the mean accuracy, with EWC, showing that P\&C achieves slightly better results avoiding catastrophic forgetting. That also happens in~\cite{rolnick2018experience}, where CLEAR method is compared with both EWC and P\&C. CLEAR outperforms EWC in most situations, and attain very similar results to P\&C.~\cite{chaudhry2019continual,ritter2018online} also use the permuted MNIST dataset and improve the mean accuracy of EWC, and some other methods they compare with. Lastly, generative strategies like~\cite{shin2017continual} prove to be efficient to prevent catastrophic forgetting. They use the datasets MNIST and SVHN to show that the performance is barely affected when they change the input dataset.

\subsection{Real Concept Drifts}
\label{sec:temporal-non-iid-behaviour}

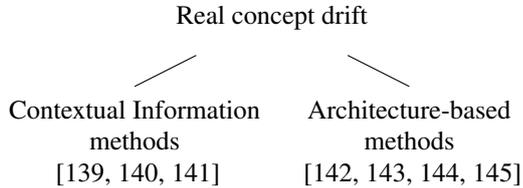
\begin{figure}[ht]
\begin{center}
\begin{forest}
for tree={
  minimum height=1cm,
  anchor=north,
  align=center,
  child anchor=north
},
[{Real concept drift},
  [{Contextual Information\\methods\\~\cite{serra2018overcoming,yoon2017lifelong,he2019task}}]
  [{Architecture-based\\methods\\~\cite{mallya2018piggyback,mallya2018packnet,masana2020ternary,rusu2016progressive}}]
]
\end{forest}
\end{center}

\caption{Classification of the different techniques able to deal with the temporal heterogeneity in the behaviour of the data samples.}
\label{fig:taxonomy-temporal-behaviours}
\end{figure}

In this kind of situation, the users are allowed to change the task they are performing during the training process. In Section~\ref{sec:spatial-behaviours} we highlighted the necessity of a model designed in a way such that different clients can have distinct outputs even though they own similar inputs. However, our interest now is pursuing a model able to flip from one output to another for the same client in different timestamps. To overcome this challenge, there are two kinds of strategies: On the one hand, we have \emph{Contextual Information Methods}, and on the other hand, we have \emph{Architecture-based Methods}, see Fig.~\ref{fig:taxonomy-temporal-behaviours}.

The approach of \emph{Contextual Information methods} is closely related to the one we talked about in Section~\ref{sec:spatial-behaviours}~\cite{yang2014unified}. That work discussed the possibility of adding a piece of information $z$ to the original data inputs, such as a task identifier or a domain identifier. This information $z$ could be designed to address both the multi-domain and the multi-task issues in a variety of situations besides the one presented in the article. Some more research articles in this line are~\cite{serra2018overcoming,yoon2017lifelong,he2019task}. The authors of~\cite{serra2018overcoming} claim the sought tasks affect the process of training, and propose using a task identifier to achieve personalization. The other approaches~\cite{yoon2017lifelong,he2019task} also rely on some kind of contextual information to determine the task corresponding to each data sample, and on the whole, they act as if they had a specific network for each of the tasks. In this case, when a new task appears in the training stage, each layer of the network is expanded and the new neurons are used for the current task, but not for previous ones.

The other kind of strategies, \emph{Architecture-based methods}, focus on deep incremental multi-task learning techniques that modify the neural network depending on the performing task, without any forgetting on previous tasks. The most studied way for accomplishing it is using some kind of mask on the neural network. Several works have explored this alternative:~\cite{mallya2018piggyback}, for instance, study the case of having a neural network already trained for one task and make use of a weight-level binary mask to cancel some of the weights, so the resulting network can solve another previously-defined task. This process can be effectively repeated for several tasks without any forgetting of the original one, as the weights are not effectively changed. There is a slightly different approach proposed by the same authors~\cite{mallya2018packnet}, which consists of starting with a neural network trained for one task, setting some of their weights to zero, i.e, eliminating some neural connections, and retraining the model a few epochs for the initial task. After that, when trying to learn a new task, the already set weights are fixed, and the eliminated connections are reestablished and trained. This method is not scalable to several tasks, as the number of neural connections is limited. Another strategy based on the same ideas uses a ternary neuron-level mask to perform training~\cite{masana2020ternary}. The reasoning behind the use of a ternary mask is that some neurons may be useful for both a new task and a previously learned one, so three possible states are considered for each neuron concerning each task: unused, used but not trainable, or trainable. This paper also faces the scalability problem, as they allow the network to grow if necessary, setting the new neurons as unused for previous tasks in order to not modify their accuracy.

On the other hand, authors of~\cite{rusu2016progressive} propose a completely different technique. They start with a deep neural network for the first task, and when they are interested in learning a new task, they start a new neural network and create connections from the ones that already existed to each layer of the new one, in order to leverage the knowledge from previous tasks. This strategy is really useful when dealing with related tasks, but tasks that interfere with each other might harm the outcome model.

\begin{table}[ht]
\centering
\resizebox{\columnwidth}{!}{%
\begin{tabular}{c|lll}
Article  & \multicolumn{3}{c}{Datasets used in experiments}                \\ \hline
\cite{serra2018overcoming}        & MNIST*;    & CIFAR-10* &     \\
\cite{yoon2017lifelong}           & MNIST*;    & CIFAR-100 &     \\
\cite{he2019task}                 & MNIST*     &           &     \\
\cite{mallya2018piggyback}        & ImagenNet; & CUBS;     & Oxford102Flowers\\
\cite{mallya2018packnet}          & ImagenNet; & CUBS;     & Oxford102Flowers\\
\cite{masana2020ternary}          & ImagenNet; & CUBS;     & Oxford102Flowers\\
\cite{rusu2016progressive}        & Atari games      \ \   & &   \\
\vphantom{a}
\end{tabular}
}
\caption{ Summary of the datasets employed in the works presented in Section~\ref{sec:temporal-non-iid-behaviour}. Asterisks indicate that the datasets have been modified in particular ways. Some of the datasets mentioned were not referenced so far: Atari games~\cite{bellemare2013arcade}  }
\label{tab:datasets-behavioursovertime}
\end{table}

These kinds of methods present a lot of differences in the way they implement their experimental results. Some of them pay attention to the accuracy obtained, while others are more concerned about the error they got, and some others concentrate on the level of forgetting they commit. For instance, in~\cite{serra2018overcoming} the authors employ the MNIST dataset with pixel permutation, like in~\cite{kirkpatrick2017overcoming}, and also exchanges some class labels in parts of the dataset to simulate the different behaviours. They compare their results with EWC, LwF~\cite{li2017learning}, and improve their results. However, they emphasize that this form of simulating the different tasks and behaviours is quite unrealistic. Surprisingly,~\cite{mallya2018piggyback,mallya2018packnet,masana2020ternary} employ the same datasets to perform their experiments: the ImageNet dataset~\cite{russakovsky2015imagenet}, used for training the pre-trained ImageNet-VGG-16 neural network; the CUBS dataset~\cite{wah2011caltech}, and the Oxford102Flowers dataset~\cite{nilsback2008automated} (see Table~\ref{tab:datasets-behavioursovertime}). This is, as we have seen in this paper, very rare.

\subsection{Data requirements for the different scenarios}
\label{sec:restrictions}

\begin{table*}[ht]
    \centering
    \includegraphics[width=0.7\textwidth]{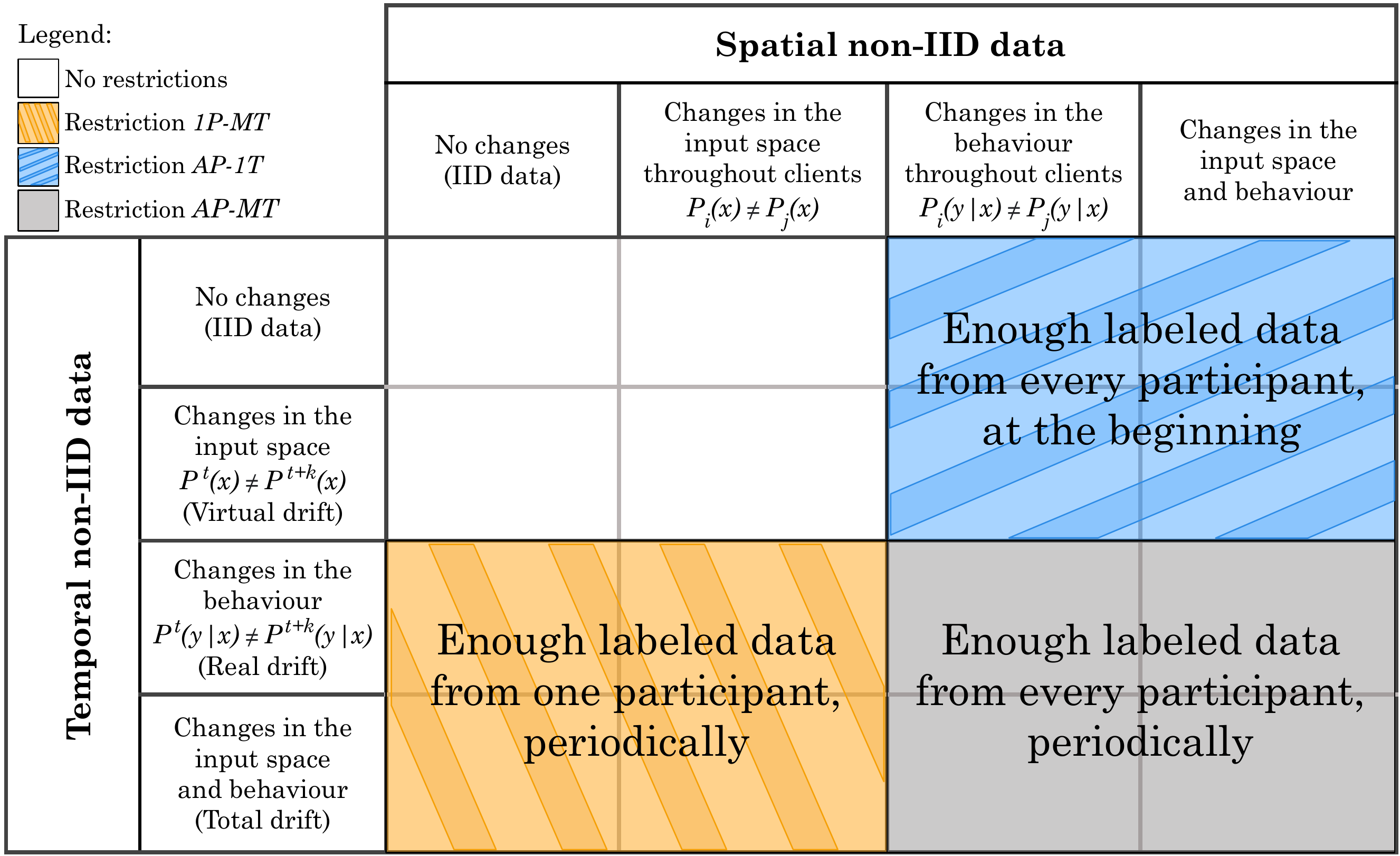}
    \caption{Required restrictions for the non-IID learning scenarios.}
    \label{tab:tabla-restricciones}
\end{table*}

To be able to apply federated learning in the different scenarios depicted in Table 4, data has to fulfill certain requirements. In this subsection we will describe these restrictions, which are summarized in Table 7.

Considering all of the scenarios presented in Table~\ref{tab:tabla-cambios}, some of them are solvable using sophisticated techniques without imposing additional restrictions, but some others may need to verify certain conditions that neither standard FL nor CL demand. As we saw in Sections~\ref{sec:federated-non-iid} and~\ref{sec:adressing-non-iid-data}, facing variations in the marginal input probabilities $\mathit{P}(x)$, either in the spatial or temporal dimension, is possible without any supplementary information~\cite{zhao2019learning,wang2010metric}, i.e, unsupervised learning techniques can also be useful in these scenarios. On the contrary, if we seek to detect changes in the conditional probability $\mathit{P}(y | x)$, a certain amount of labeled data is required~\cite{yang2014unified,mallya2018piggyback}, because these kinds of changes can only be measured with the error committed. To be more precise, we establish three restrictions that need to be satisfied to face some scenarios, and denote them as Restrictions 1P-MT, AP-1T, and AP-MT in Table~\ref{tab:tabla-restricciones}:

\begin{itemize}
    \item[i)] If the clients behaviour change with time but does not change among the devices, i.e., data fit the cells marked as \emph{Restriction 1P-MT} (one participant, many times) in Table~\ref{tab:tabla-restricciones}, then we will need to have enough labeled data of at least one participant from time to time. Knowing the behaviour of one participant is enough since in these scenarios the behaviours of all of the other participants will be the same. When a Real Concept Drift occurs, that client labeled data will allow the model to detect that drift and properly react to it.
    \item[ii)] If, on the contrary, data fit the cells marked as \emph{Restriction AP-1T} (all participants, one time) in Table~\ref{tab:tabla-restricciones}, then the clients are allowed to present different conditional probabilities, but they will remain constant in time. In that situation, enough labeled data from all of the participants will be required at the beginning of the training process, so we can determine their behaviour. Once their behaviour is settled, it is not possible for it to change, so no more labeled data is required.
    \item[iii)]  Lastly, if conditional probabilities vary both in the spatial and temporal axis, which corresponds to cells marked as \emph{Restriction AP-MT} (all participants, many times) in Table~\ref{tab:tabla-restricciones}, then we need enough labeled data from all of the participants, from time to time, so we can conclude when drifts occur and act in consequence. This restriction provides strictly more information than the other ones, so any other scenario considered in Table~\ref{tab:tabla-restricciones} will also be solvable under this requirement. However, it could be very unrealistic to assume that we could have this information in real-world problems.
\end{itemize}

The kind of heterogeneity that can be handled without any additional restriction is by far the most studied one in the literature (see the number of works cited in Table~\ref{tab:tabla-cambios}). Situations where some additional condition is required are less studied, not because the tasks that fit these scenarios are uncommon, but because it is harder to work under the restrictions we just settled. 

\section{Challenges and future directions}
\label{sec:challenges}

Along this review, we have discussed the different possible causes of data heterogeneity, as well as the most common and remarkable strategies developed so far to face it. Some of those strategies are already designed and implemented in the FL framework, while some others seem promising, but at the moment are only conceived in centralized settings. At the same time, we have addressed the necessity of considering time-evolving methods for real-life federated problems. Some works are aware of this kind of issue, but nowadays this area of research is much less studied than the one regarding non-IID data.

Concerning the non-IID data, there are some important existing problems. One of the biggest ones is that most of the strategies designed to tackle non-IID data do not specify what kind of non-IID data source they work with. See, for instance, works presented in Sections~\ref{sec:personalization} and \ref{sec:strategies-for-non-iid}. Hence, when trying to apply some method for a real-life problem, it is unclear to determine which ones are useful, or if some of them are more appropriate than others. Also, the fact that a lots of works claim to deal with non-IID data leads to the thinking that there are a lot of different techniques in the current literature to solve non-IID data problems, but the reality is that some kinds of heterogeneity are still barely studied.

Nowadays, personalization strategies (Section~\ref{sec:personalization}) are gaining a lot of importance in FL. These methods are aware of the possibility of having clients that need their own outputs for their data. On the contrary, most of the current strategies in the literature assume that the same input data belonging to different participants must be replied to with the same output. As we saw in Sections~\ref{sec:spatial-behaviours} and~\ref{sec:temporal-non-iid-behaviour}, this is not always true.

Moreover, some of the proposed techniques for handling spatial non-IID data are already conceived in a FL framework, such as Generative Adversarial Networks. Nonetheless, some strategies are yet to be deployed in FL settings. Is the case of Domain Factorization methods and Dissimilarity methods. Regarding Domain Factorization, the main challenge when trying to perform these strategies in a federated setting is that each participant would construct a different input space factorization, and it is necessary to establish a common ground for all of them.  Concerning Dissimilarity methods, the main challenge is establishing a common metric that generalizes the domain variability of all participants.
 
Concerning the temporal dimension, it is important to notice that the current strategies of drift detection and adaptation are mostly deployed in centralized settings. However, we selected the strategies they employ because they could be easily adapted to federated settings. For instance, in a federated environment, each of the participants could perform a rehearsal technique to avoid forgetting previous concepts. Another possibility, considering Regularization Methods, is that clients who perceive different domains train particular neurons of the network, leading to a faster domain adaptation. Nonetheless, some difficulties can arise in these situations:

\begin{itemize}
    \item Clients may experience drifts at different timestamps, and thus they present different input domains simultaneously. This can lead to very different updates for the global model, and prevent the global model from converging.
    \item Clients may experience similar drifts in their data without being aware of it, and mechanisms that provide this information would facilitate achieving a better model faster. 
\end{itemize}

In addition to the inherent difficulties of considering heterogeneous data, the lack of specific datasets makes it harder to test the quality of the methods under these settings. At the present moment, each work employs different datasets to test their methods, and in the majority of cases, they need to modify those datasets to create the desired data distributions (see Tables~\ref{tab:datasets-personalization},\ref{tab:datasets-inputspacethroughclients},\ref{tab:datasets-inputspaceovertime} and \ref{tab:datasets-behavioursovertime}). Under these circumstances, it is impossible to fairly compare those methods. It is necessary to have a common benchmark dataset, with standard representations of some types of heterogeneous data. This would allow to contrast the current and future strategies over a common set of data and properly compare them.

\section*{Acknowledgements}

This work has received financial support from AEI/FEDER~(EU) grant number PID2020-119367RB-I00. It has also been supported by the Xunta de Galicia - Consellería de Cultura, Educación e Universidade (Centros de investigación de Galicia accreditation 2019-2022 ED431G-2019/04 and ED431G2019/01, and Reference Competitive Groups accreditation 2021-2024, ED431C 2018/29, ED431F2018/02 and ED431C 2021/30) and the European Union (European Regional Development Fund~-~ERDF). Finally, it has also been funded by the Ministerio de Universidades of Spain in the FPU 2017 program (FPU17/04154).

%
% ---- Bibliography ----
%
\bibliographystyle{elsarticle-num}
\bibliography{main}

\end{document}